\pdfoutput=1

\documentclass[11pt]{article}

\usepackage[final]{coling}

\usepackage{times}
\usepackage{latexsym}
\usepackage{amssymb}
\usepackage{amsmath}
\usepackage{makecell}
\usepackage{bm}
\usepackage{booktabs}
\usepackage{multirow}
\usepackage{float}
\makeatletter
\renewcommand{\maketag@@@}[1]{\hbox{\m@th\normalsize\normalfont#1}}%
\makeatother

\usepackage[T1]{fontenc}

\usepackage[utf8]{inputenc}

\usepackage{microtype}

\usepackage{inconsolata}

\usepackage{graphicx}

\title{Reasoning-Oriented and Analogy-Based Methods for Locating and Editing in Zero-Shot Event-Relational Reasoning}

\author{
  \textbf{Jingyao Tang \textsuperscript{1}}
  \textbf{Lishuang Li \textsuperscript{1}}\thanks{Corresponding author}
  \textbf{Liteng Mi \textsuperscript{1}}
  \textbf{Haiming Wu \textsuperscript{2}}
  \textbf{Hongbin Lu \textsuperscript{3}} \\
  \textsuperscript{1}School of Computer Science and Technology, Dalian University of Technology\\
  \textsuperscript{2}School of Computer Science and Technology, Beijing Institute of Technology\\
  \textsuperscript{3}School of Computer Science and Artificial Intelligence, Liaoning Normal University\\
  \href{mailto:tangjingyao@mail.dlut.edu.cn}{tangjingyao@mail.dlut.edu.cn}, \href{mailto:lils@dlut.edu.cn}{lils@dlut.edu.cn}, \href{mailto:dlutmlt@mail.dlut.edu.cn}{dlutmlt@mail.dlut.edu.cn}\\
  \href{mailto:hm.wu@bit.edu.cn}{hm.wu@bit.edu.cn}, \href{mailto:luhongbin-123@163.com}{luhongbin-123@163.com}
}

\begin{document}
\maketitle
\begin{abstract}
Zero-shot event-relational reasoning is an important task in natural language processing, and existing methods jointly learn a variety of event-relational prefixes and inference-form prefixes to achieve such tasks. However, training prefixes consumes large computational resources and lacks interpretability. Additionally, learning various relational and inferential knowledge inefficiently exploits the connections between tasks. Therefore, we first propose a method for Reasoning-Oriented Locating and Editing (ROLE)\footnote{\href{https://github.com/manderous/ROLE_ABLE}{https://github.com/manderous/ROLE\_ABLE}\label{MyGit}}, which locates and edits the key modules of the language model for reasoning about event relations, enhancing interpretability and also resource-efficiently optimizing the reasoning ability. Subsequently, we propose a method for Analogy-Based Locating and Editing (ABLE)\textsuperscript{\ref{MyGit}}, which efficiently exploits the similarities and differences between tasks to optimize the zero-shot reasoning capability. Experimental results show that ROLE improves interpretability and reasoning performance with reduced computational cost. ABLE achieves SOTA results in zero-shot reasoning.
\end{abstract}

\section{Introduction}

In the information extraction domain, reasoning about relations (e.g., causal, temporal, sub-events) between events \cite{man2024mastering, niu2024contempo, wang2022maven, lai2022multilingual} is crucial. These relationships have been used to construct event graphs \cite{frisoni2022text, chen2022ergo}, event prediction \cite{shi2024language}, commonsense reasoning \cite{lv2024disentangled}, dialog generation \cite{wang2024research}, and question answering \cite{majumdar2024openeqa}.

\begin{figure}[t]
  \begin{center}
    \includegraphics[width=0.8\columnwidth]{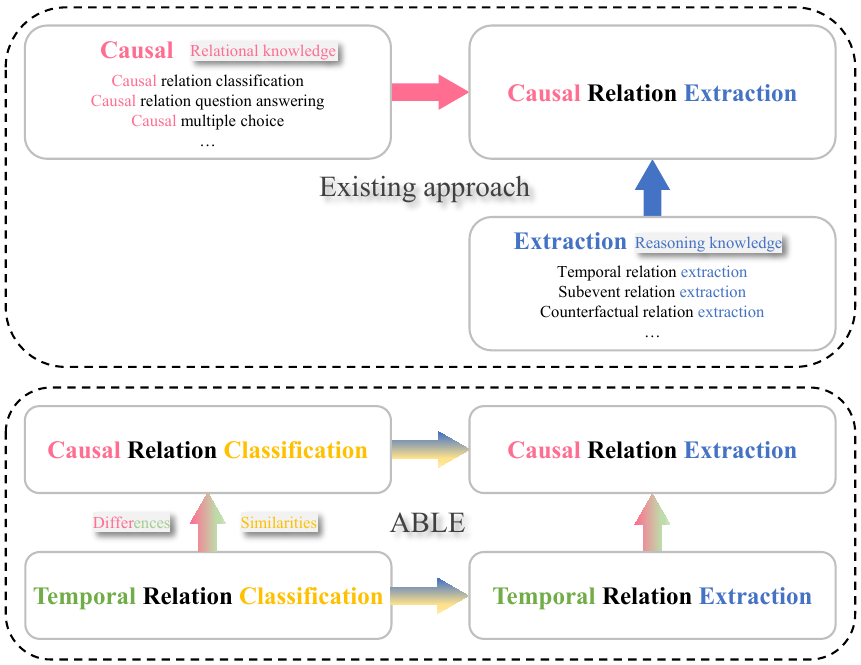}
    \caption{Comparison of knowledge transfer between existing methods and ABLE in various event-relational reasoning tasks. Top: Existing methods rely on many tasks to learn relational and reasoning knowledge, inefficiently exploiting the connections between tasks. Bottom: ABLE efficiently learns similarities and differences between tasks to enhance knowledge transfer.}
    \label{Fig1}
  \end{center}
\end{figure}

Due to the limitations of manual labeling, we turn our attention to zero-shot event-relational reasoning. Existing approaches \cite{tao2023unievent} use a multi-task framework to jointly learn the various relational and inferential prefixes, and then use the corresponding prefixes to achieve zero-shot relational reasoning. However, fine-tuning prefixes requires high computational cost and lacks interpretability. In addition, learning multiple relational and inferential knowledge inefficiently utilizes connections between tasks (see Figure~\ref{Fig1}).

Therefore, we propose a method for Reasoning-Oriented Locating and Editing (ROLE), which locates the key modules of the language model in event-relational reasoning, and explores the reasoning mechanism, thus enhancing the interpretability. Meanwhile, ROLE edits the key modules, resource-efficiently optimizing the reasoning ability of the language model. Moreover, we propose a method for Analogy-Based Locating and Editing (ABLE), which learns the similarities and differences among various tasks, and efficiently migrates knowledge, thereby enhancing zero-shot reasoning (see Figure~\ref{Fig1}).

We locate key modules on 6 event-relational reasoning tasks and evaluate the performance of reasoning about event relations on 10 datasets. The experimental results show that ROLE improves interpretability and reasoning performance with reduced computational cost; ABLE achieves State-Of-The-Art (SOTA) results of zero-shot event-relational reasoning on most datasets. Our main contributions are as follows:

(1) We propose ROLE, which locates and edits key modules for reasoning about event relations in language models, and explores reasoning mechanisms, further improving interpretability and reasoning performance with reduced computational cost.

(2) We propose ABLE, which exploits similarities and differences in analogies among tasks, thereby achieving SOTA results for zero-shot reasoning. Additionally, we analyze the analogicality of locating and editing to further validate the effectiveness of ABLE.

\section{Related Work}

\subsection{Zero-shot event-relational reasoning}

As Large Language Models (LLMs) rapidly evolve, the reasoning ability of LLMs have drawn attention. Tao et al. \cite{tao2023unievent} proposed UniEvent, which prefix-tune T5 to optimize zero-shot event-relational reasoning. Moreover, Yuan et al. \cite{yuan2023zero} investigated the performance of GPT-3.5 in zero-shot temporal-relational extraction, and Gao et al. \cite{gao2023chatgpt} evaluated the ability of various LLMs in causal reasoning tasks. Subsequently, Tao et al. \cite{tao2024comprehensive} provided zero-shot results for event-relational reasoning using different LLMs.

These studies suggest that LLMs are competitive in relational reasoning tasks. However, the underlying reasoning mechanisms in LLMs remain underexplored. Therefore, our approach aims to explore the key modules and reasoning mechanisms of language models from an interpretable perspective.

\subsection{Knowledge editing}

Existing methods can be divided into two categories \cite{yao2023editing}: preserving the parameters of the model, and modifying them directly. The first type of methods \cite{huang2023transformer, dong2022calibrating} adds additional parameters to update the knowledge without changing the original parameters. The second type of methods \cite{mitchell2021fast, meng2022locating, meng2022mass} directly updates the internal parameters of the model. For example, MEMIT \cite{meng2022mass} achieves precise modification of knowledge by locating key modules and editing them, and are effective in reducing computational cost.

These approaches aim to insert or update knowledge by adjusting the parameters of specific modules without re-training the model. This motivated us to consider whether there are critical modules that can be edited to improve the reasoning ability.

\section{Method}

\begin{figure*}[t]
  \begin{center}
    \includegraphics[width=0.8\linewidth]{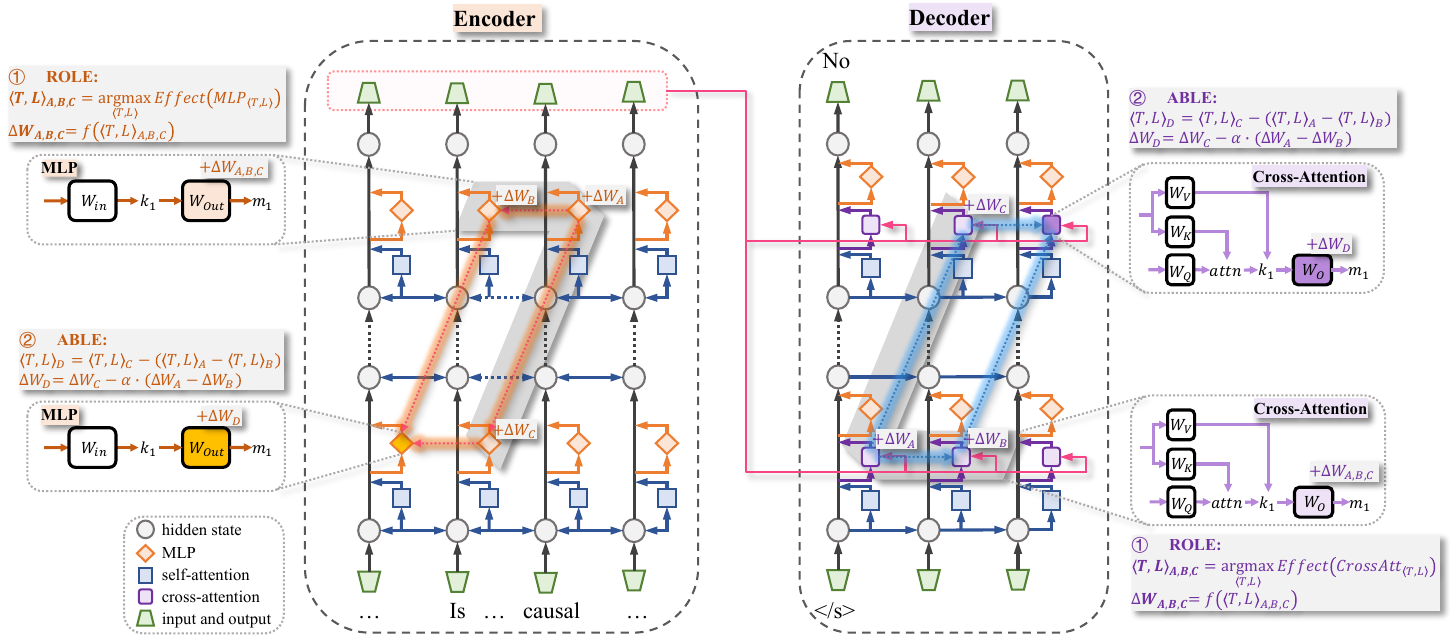}
    \caption{Overview of ROLE and ABLE. The left side shows the application of ROLE and ABLE in the Flan-T5-large encoder, and the right side shows their application in the decoder. First, ROLE is used to determine the position and editing information of tasks \begin{math} A \end{math}, \begin{math} B \end{math} and \begin{math} C \end{math} (corresponding to the three vertices of the parallelogram). Then, ABLE migrates this information to task \begin{math} D \end{math} (the fourth vertex) by analogy.}
    \label{Fig2}
  \end{center}
\end{figure*}

Recently, researchers have started to utilize generative models (e.g., T5) for event-relational reasoning \cite{man2022event, man2024hierarchical, chen2024event, yang2024temprompt}, as such models utilize prompts more efficiently. Therefore, in this section, we propose ROLE and ABLE using Flan-t5-large as the backbone model, and their overall framework is shown in Figure ~\ref{Fig2}.

\subsection{Reasoning-oriented locating and editing}

We propose reasoning-oriented locating and editing, inspired by knowledge editing \cite{meng2022mass}. First, reasoning-oriented locating identifies the key modules \begin{math} H_{\langle{T,L}\rangle} \end{math} of the language model in the reasoning task. Second, reasoning-oriented editing computes the change magnitude \begin{math} {\mathrm{\Delta}W}_{H_{\langle{T,L}\rangle}} \end{math} of the key module to optimize the reasoning performance.

\subsubsection{Reasoning-oriented locating}

This subsection aims to identify key modules. We iterate over each module \begin{math} 
h_{\langle{t,l}\rangle} \end{math} of the language model and compute the effect of these modules on the reasoning task using the average indirect effect \cite{pearl2022direct}. For positive samples, the formula for calculating the effect is:
\vspace{-5pt}
\begin{equation}
\resizebox{0.87\hsize}{!}{$Effect\left( h_{\langle{t,l}\rangle} \right) = {\sum_{x_{pos}}{\left\lbrack {P\left( Yes \middle| x_{pos}^{*},h_{\langle{t,l}\rangle} \right) - P\left( Yes \middle| x_{pos}^{*},h_{\langle{t,l}\rangle}^{*} \right)} \right\rbrack,}}$} \label{eq1}
\end{equation}

For negative samples, the effect is given by:
\vspace{-5pt}
\begin{equation}
\resizebox{0.87\hsize}{!}{$
Effect\left( h_{\langle{t,l}\rangle} \right) = {\sum_{x_{neg}}{\left\lbrack {P\left( No \middle| x_{neg}^{*},h_{\langle{t,l}\rangle}^{*} \right) - P\left( No \middle| x_{neg}^{*},h_{\langle{t,l}\rangle} \right)} \right\rbrack,}}$} \label{eq2}
\end{equation}
where, \begin{math} h \end{math} denotes the module type in the language model, \begin{math} t \end{math} and \begin{math} l \end{math} denote the token and the layer number, respectively, so that \begin{math} h_{\langle{t,l}\rangle} \end{math} denotes the module \begin{math} h \end{math} in the \begin{math} l \end{math}-th layer associated with token \begin{math} t \end{math}. In addition, \begin{math} x \end{math} is the prompt, \begin{math} 
x^{*} \end{math} is the prompt with noise, and \begin{math} 
h^{*} \end{math} denotes the noise-affected module. The reverse order of the subtraction of conditional probabilities in Eq.~(\ref{eq2}) and Eq.~(\ref{eq1}) is to ensure that the value of the \begin{math} Effect \end{math} is positive. Because, for negative samples, noise disrupts the key information in the prompts, making it difficult for the model to accurately capture the relation type, i.e., the model prefers to output "No" under the condition of \begin{math} 
h^{*} \end{math}.

Thus, we first select the key module type \begin{math} H \end{math} based on the overall Effect. Then identify the position \begin{math} \left\langle {T,L} \right\rangle \end{math} with the largest Effect to determine the key module \begin{math} H_{\langle{T,L}\rangle} \end{math}:

\vspace{-3pt}

\begin{small}
\begin{equation}
\left\langle {T,L} \right\rangle = {\underset{\langle{t,l}\rangle}{argmax}{Effect\left( H_{\langle{t,l}\rangle} \right)}}. \label{eq3}
\end{equation}
\end{small}

Specifically, we study 4 module types, including Transformer, MLP, Self-Attention, and Cross-Attention (available only for Decoder) modules. Meanwhile, we focus on 6 types of reasoning tasks, including classification and extraction tasks for temporal, causal and sub-event relations. The design of the prompt and verbalizers, the setup of positive and negative samples, and other preprocessing specifics are detailed in Appendix~\ref{appendix_A} as well as in Table~\ref{tab10}.

\begin{figure*}[t]
  \begin{center}
    \includegraphics[width=0.375\linewidth]{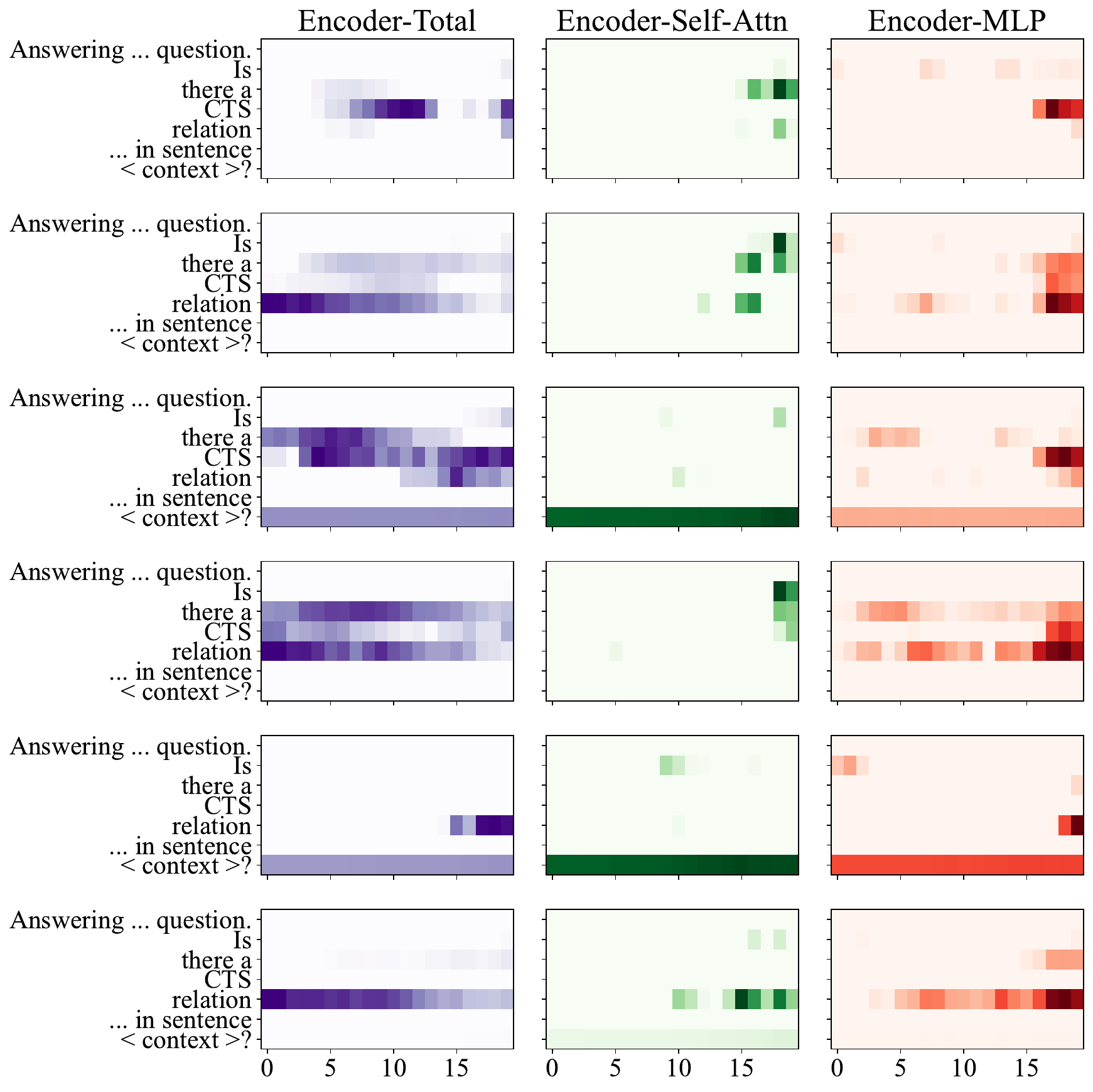} \hspace{3mm}
    \includegraphics[width=0.3\linewidth]{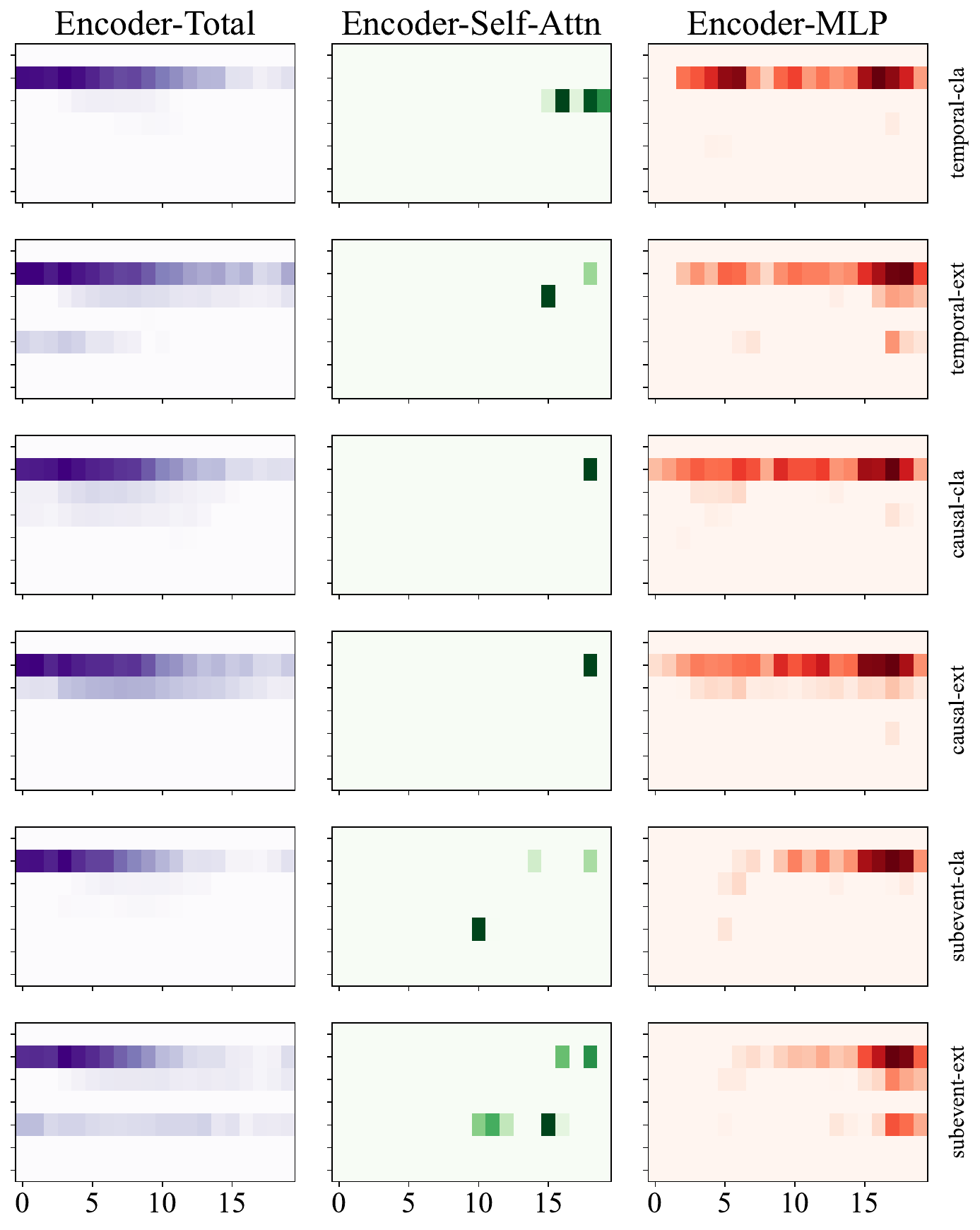}
    \caption {Heatmaps of the effect of Transformer, MLP, and Self-Attention modules on each token for each layer in the encoder, with positive samples on the left and negative samples on the right. The horizontal axis indicates the number of layers, the vertical axis indicates the tokens, and the color depth indicates the intensity of the effect. tokens are presented in 7 groups (see Appendix~\ref{appendix_C} for details).}
    \label{Fig3}
  \end{center}
\end{figure*}

\begin{figure*}[t]
  \begin{center}
    \includegraphics[width=0.38\linewidth]{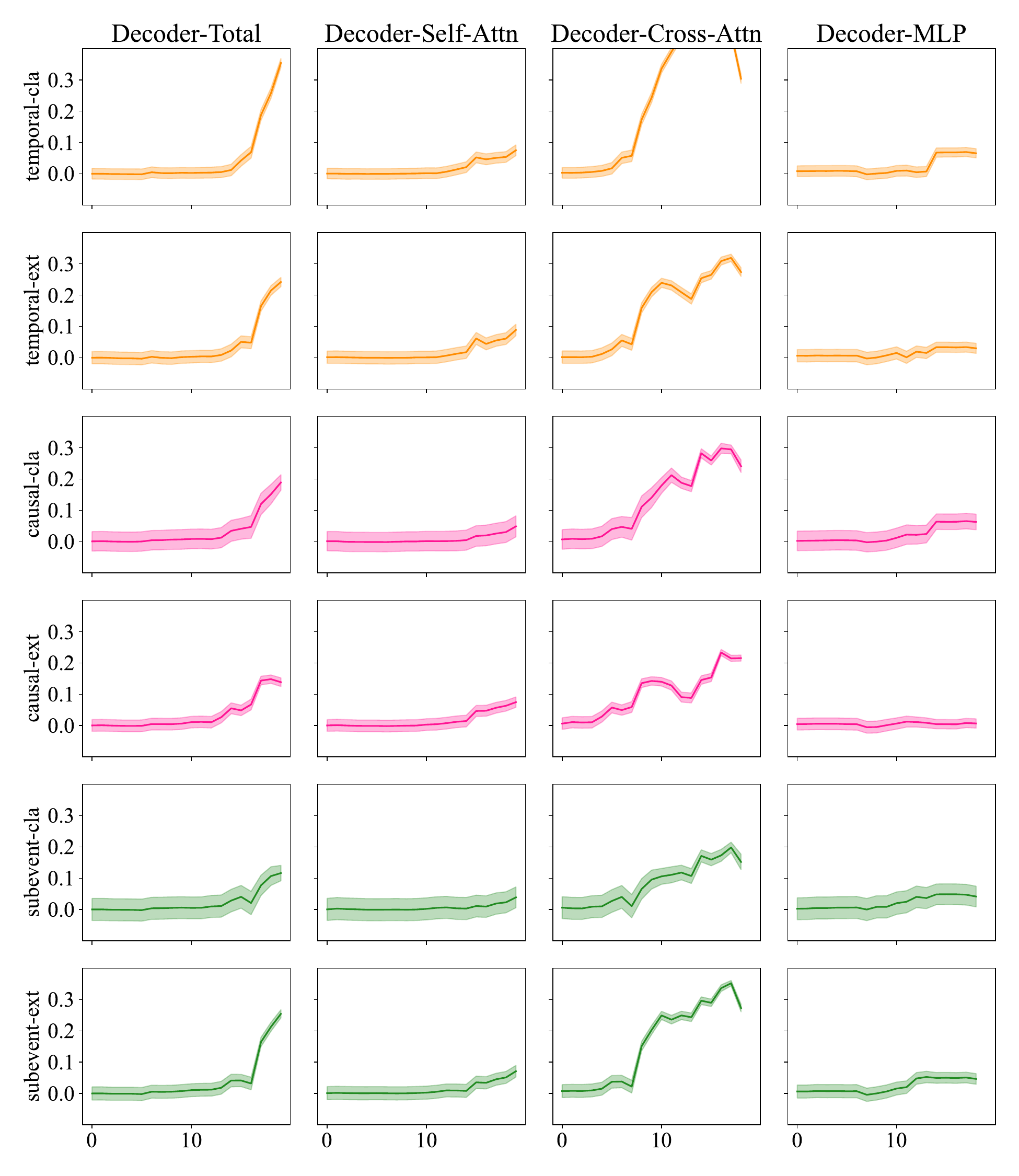}  \hspace{-0.1mm}
    \includegraphics[width=0.35\linewidth]{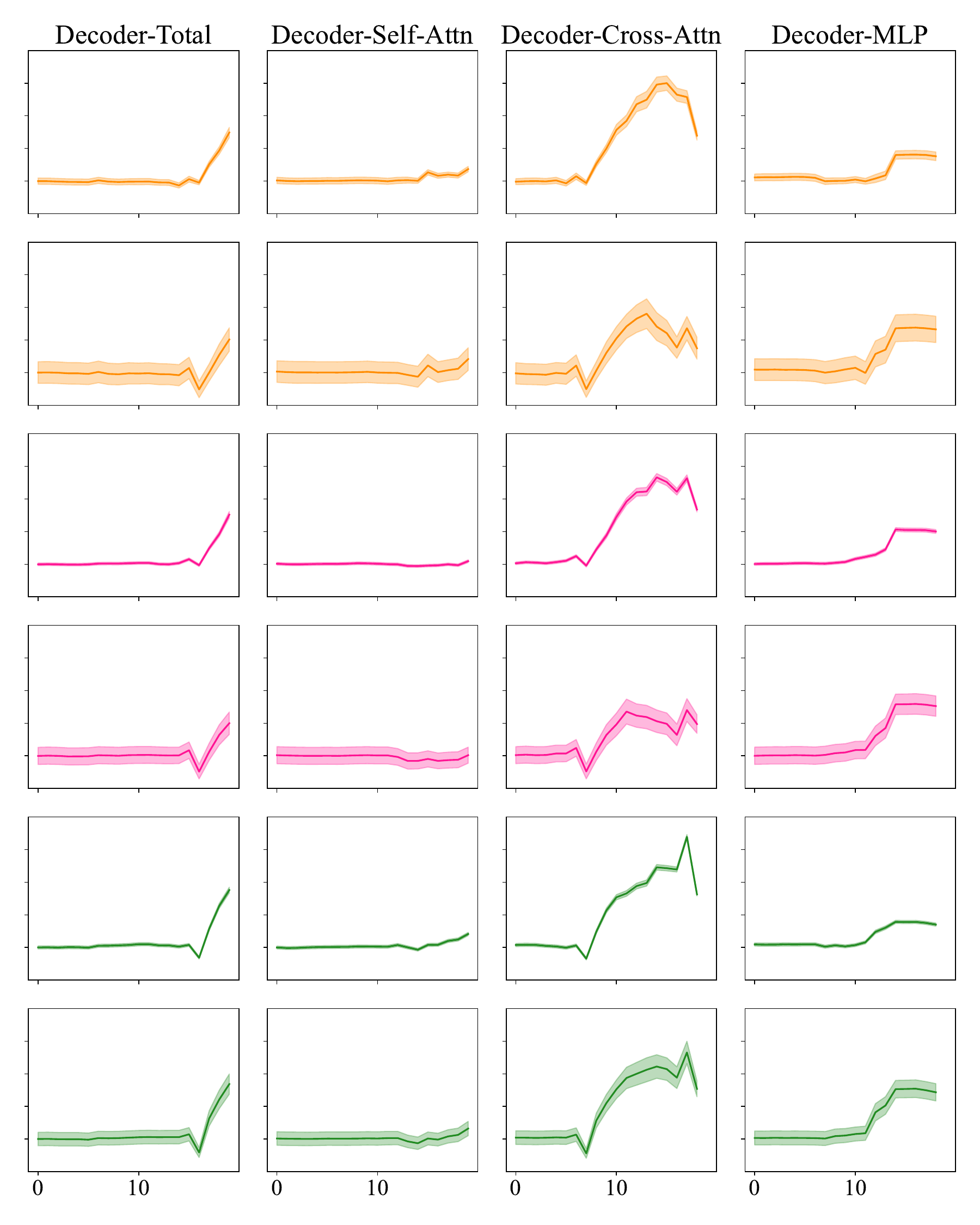}
    \caption {Line plots of the effect of the Transformer, MLP, Self-Attention and Cross-Attention modules on the </s> token for each layer in the decoder, with positive samples on the left and negative samples on the right. The horizontal axis indicates the number of layers and the vertical axis indicates the intensity of the impact.}
    \label{Fig4}
  \end{center}
\end{figure*}

\begin{figure}
  \begin{center}
    \includegraphics[width=0.8\columnwidth]{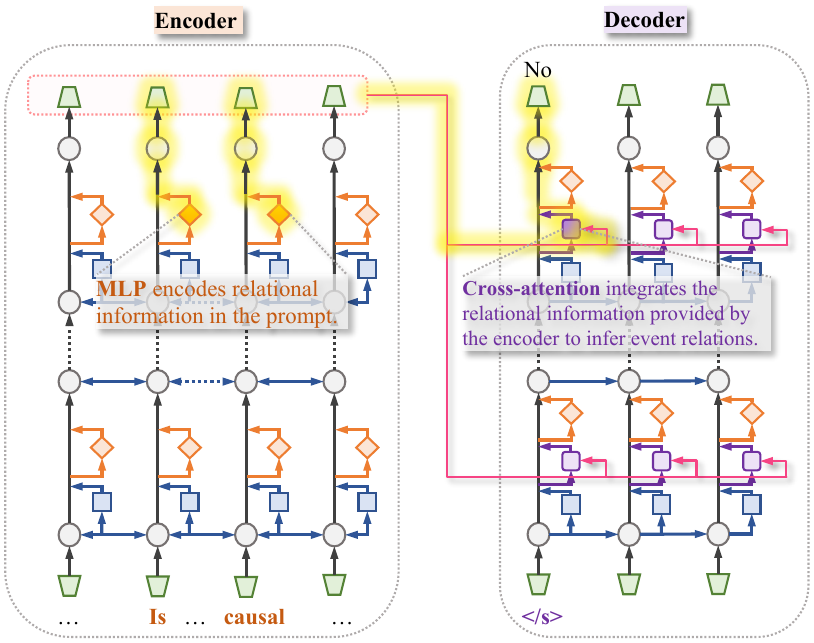}
    \caption{Reasoning mechanism for Flan-T5-large inferring event relations.}
    \label{Fig5}
  \end{center}
\end{figure}

Figure~\ref{Fig3} and Figure~\ref{Fig4} show that the MLP module in the encoder and the Cross-Attention module in the decoder have the greatest overall impact on all reasoning tasks, as their impact graphs are most similar to those of the corresponding Transformer. Further, Table~\ref{tab1} shows the locations where these two types of modules have the greatest impact on different tasks, leading to key modules \begin{math} H_{\langle{T,L}\rangle} \end{math} being identified. Finally, we analyze the location of key modules and propose the following hypothesis for the reasoning mechanism:

\textbf{Hypothesis (Reasoning Mechanism for Event-Relational Reasoning Task in Flan-t5-large)}: In the encoder, the MLP module encodes relational information in the prompts, including relation-type words (e.g., ``causal'' and ``relatio'') and question words (e.g., ``Is''). In the decoder, the cross-attention module integrates the relational information provided by the encoder with the start token (e.g., ``</s>'') to infer event relations, as shown in Figure~\ref{Fig5}.

\begin{table}
\resizebox{3in}{!}{
\begin{tabular}{lcccc}
\hline
& \multicolumn{2}{c}{Encoder's MLP} & \multicolumn{2}{c}{Decoder's cross-attention} \\
\hline
& Positive & Negative & Positive & Negative \\
\hline
Temporal   relation classification  & \textless{}”temporal”, 17\textgreater{} & \textless{}”Is”, 16\textgreater{} & \textless{}”\textless{}/s\textgreater{}”, 14\textgreater{} & \textless{}”\textless{}/s\textgreater{}”, 15\textgreater{} \\
Temporal   relation extraction      & \textless{}”relation”, 17\textgreater{} & \textless{}”Is”, 18\textgreater{} & \textless{}”\textless{}/s\textgreater{}”, 17\textgreater{} & \textless{}”\textless{}/s\textgreater{}”, 13\textgreater{} \\
Causal relation   classification    & \textless{}”causal”, 18\textgreater{}   & \textless{}”Is”, 17\textgreater{} & \textless{}”\textless{}/s\textgreater{}”, 16\textgreater{} & \textless{}”\textless{}/s\textgreater{}”, 14\textgreater{} \\
Causal relation   extraction        & \textless{}”relation”, 18\textgreater{} & \textless{}”Is”, 17\textgreater{} & \textless{}”\textless{}/s\textgreater{}”, 16\textgreater{} & \textless{}”\textless{}/s\textgreater{}”, 17\textgreater{} \\
Sub-event   relation classification & \textless{}”relation”, 19\textgreater{} & \textless{}”Is”, 17\textgreater{} & \textless{}”\textless{}/s\textgreater{}”, 17\textgreater{} & \textless{}”\textless{}/s\textgreater{}”, 17\textgreater{} \\
Sub-event   relation extraction     & \textless{}”relation”, 18\textgreater{} & \textless{}”Is”, 17\textgreater{} & \textless{}”\textless{}/s\textgreater{}”, 17\textgreater{} & \textless{}”\textless{}/s\textgreater{}”, 17\textgreater{} \\
\hline
\end{tabular}}
\caption{Locations (both tokens and layers) of the encoder's MLP module and the decoder's cross-attention module that most significantly affect performance across tasks. ``Positive'' denotes positive samples and ``Negative'' denotes negative samples. Refer to Table~\ref{tab10} and Table~\ref{tab11} in Appendix~\ref{appendix_B} for detailed layer ordering.}
\label{tab1}
\end{table}

\subsubsection{Reasoning-oriented editing}

This subsection aims to compute the magnitude of editing \begin{math} {\mathrm{\Delta}W}_{H_{\langle{T,L}\rangle}} \end{math} in the parameters of the localized module \begin{math} H_{\langle{T,L}\rangle} \end{math}. We edit the weights \begin{math} 
W_{out} \end{math} and \begin{math} 
W_{O} \end{math} of the last linear layer in the MLP module and the cross-attention module, respectively (see Figure~\ref{Fig2}), because they directly affect the output of the modules and modifying them can most effectively influence the model's decisions.

Moreover, we find that the T5 model tends to answer “Yes” when inferring event relations (recall is much higher than precision as observed in Table~\ref{tab3} and Table~\ref{tab5}). This tendency aligns with existing research, which, for example, shows that LLMs are inclined to identify events as causally related \cite{gao2023chatgpt}. This inclination arises from the “memory hallucination” \cite{mckenna2023sources}, as the pre-training corpus contains a large number of causally related samples, and thus the model may tend to judge the test samples as causally related as well. To mitigate this tendency, we construct an objective function for the output of the module \begin{math} H_{\langle{T,L}\rangle} \end{math}:

\begin{scriptsize}
\begin{equation}
\begin{aligned}
&M_{1} = \underset{m}{\arg\min}\left\lbrack 
- \sum_{x_{neg}}{\log P_{H_{\langle T,L \rangle} = m}\left( \text{No} \middle| x_{neg} \right)} \right. \\
&\left. + \sum_{x_{pos}}{D_{KL}\left\lbrack P_{H_{\langle T,L \rangle} = m}\left( \text{Yes} \middle| x_{pos} \right) 
\middle| \middle| P\left( \text{Yes} \middle| x_{pos} \right) \right\rbrack} \right\rbrack, \\
&H \in \left\{ \text{Enc}_{MLP},~\text{Dec}_{CrossAtt} \right\},
\end{aligned}
\label{eq4}
\end{equation}
\end{scriptsize}
where, \begin{math} x_{neg} \end{math} and \begin{math} x_{pos} \end{math} denote negative and positive samples, respectively. \begin{math} 
P_{H_{\langle{T,L}\rangle} = m}( \cdot ) \end{math} denotes the output probability of the model after updating the output of the module \begin{math} H_{\langle{T,L}\rangle} \end{math} to \begin{math} m \end{math}. \begin{math} 
M_{1} \end{math} the updated output of \begin{math} H_{\langle{T,L}\rangle} \end{math}. \begin{math} D_{KL}\lbrack \cdot \rbrack \end{math} computes the KL divergence. This objective function aims to let the false negative samples answer “No” after editing, while keeping the positive samples still answering “Yes”. Thus, according to the theory proposed by Meng et al. \cite{meng2022mass}, using the updated output \begin{math} 
M_{1} \end{math}, we can obtain the editing magnitude \begin{math} 
{\mathrm{\Delta}W}_{H_{\langle{T,L}\rangle}} \end{math} of the module \begin{math} H_{\langle{T,L}\rangle} \end{math}:

\vspace{-5pt}

\begin{small}
\begin{equation}
\resizebox{0.7\hsize}{!}{${\mathrm{\Delta}W}_{H_{\langle{T,L}\rangle}} = RK_{1}^{T}\left( {C_{0} + K_{1}K_{1}^{T}} \right)^{- 1},$} \label{eq5}
\end{equation}
\end{small}
where, \begin{math} 
R \triangleq M_{1} - W_{0}K_{1} \end{math},\begin{math} C_{0} = \lambda \cdot E_{k}\left\lbrack {kk^{T}} \right\rbrack \end{math}. \begin{math} 
K_{1} \end{math} denotes the input of the module \begin{math} H_{\langle{T,L}\rangle} \end{math} , which can be computed by forward propagation. \begin{math} 
W_{0} \end{math} denotes the original parameters of the module \begin{math} H_{\langle{T,L}\rangle} \end{math}. We used the corpus Colossal Clean Crawled Corpus (C4) \cite{raffel2020exploring} from pre-training T5 to compute \begin{math} k \end{math} and \begin{math} 
C_{0} \end{math}. \begin{math} \lambda \end{math} is a hyperparameter.

\subsection{Analogy-based locating and editing approach}

This subsection aims to fully utilize the similarities and differences between tasks to enhance zero-shot reasoning. We set four analogizable tasks \begin{math} A \end{math}, \begin{math} B \end{math}, \begin{math} C \end{math} and \begin{math} D \end{math}, i.e., the relationship between task \begin{math} A \end{math} and task \begin{math} B \end{math} can be analogized to the relationship between task \begin{math} C \end{math} and task \begin{math} D \end{math}. We first utilize ROLE to obtain the locating and editing information of tasks \begin{math} A \end{math}, \begin{math} B \end{math} and \begin{math} C \end{math}:

\begin{scriptsize}
\begin{equation}
\begin{aligned}
&\left\langle {T,L} \right\rangle_{Task} = {\underset{\langle{T,L}\rangle}{argmax}\left\lbrack {P\left( No \middle| x^{*},~H_{T,L}^{*} \right) - P\left( No \middle| x^{*},H_{T,L} \right)} \right\rbrack}, \\
&~H \in \left\{ {Enc_{MLP},~Dec_{CrossAtt}} \right\},~Task \in \left\{ {A,B,C} \right\}, \label{eq6}
\end{aligned}
\end{equation}
\end{scriptsize}

\vspace{-15pt}

\begin{equation}
\resizebox{0.85\hsize}{!}{${\mathrm{\Delta}W}_{Task} = RK_{1}^{T}\left( {C_{0} + K_{1}K_{1}^{T}} \right)^{- 1},~Task \in \left\{ {A,B,C} \right\},$} \label{eq7}
\end{equation}
where, \begin{math} 
{\mathrm{\Delta}W}_{h_{\langle{T_{1}L_{1}}\rangle}} \end{math} simplifies to \begin{math} \mathrm{\Delta}W \end{math}. These informations are then migrated analogously to task D (see Figure~\ref{Fig2}):

\begin{equation}
\resizebox{0.7\hsize}{!}{$\left\langle {T,L} \right\rangle_{D} = \left\langle {T,L} \right\rangle_{C} - \left( {\left\langle {T,L} \right\rangle_{A} - \left\langle {T,L} \right\rangle_{B}} \right),$} \label{eq8}
\end{equation}

\vspace{-15pt}

\begin{equation}
\resizebox{0.7\hsize}{!}{$
{\mathrm{\Delta}W}_{D} = {\mathrm{\Delta}W}_{C} - \alpha \cdot \left( {{\mathrm{\Delta}W}_{A} - {\mathrm{\Delta}W}_{B}} \right),$} \label{eq9}
\end{equation}
where, \begin{math} \alpha \end{math} is a hyperparameter that regulates the degree of being analogized. Finally, we optimize zero-shot learning using the locating and editing information of task \begin{math} 
D \end{math}.

\section{Experiment}

\subsection{Datasets}

We perform zero-shot event-relational reasoning tasks on 10 datasets, covering three types of tasks: causal relation extraction, causal relation classification, and sub-event relation extraction.

Causal relation extraction: following Gao et al. \cite{gao2023chatgpt}, we evaluate intra-sentence pairs of causal events in EventStoryLine v0.9 (\textbf{ESC-intra}) \cite{caselli2017event}, Causal-TimeBank (\textbf{CTB-intra}) \cite{mirza2014annotating} and MAVEN-ERE (\textbf{MAVEN-intra-causal)} \cite{wang2022maven}. Furthermore, following the work of Tao et al. \cite{tao2023unievent} for UniEvent, we evaluate SCITE (\textbf{SCI-uni}), EventStoryLine (\textbf{ESL-uni}) and Causal-TimeBank (\textbf{CTB-uni}).

Causal relation classification: we evaluate Causal News Corpus (\textbf{CNC}) \cite{tan2022causal} and AltLex (\textbf{ALT-uni}) \cite{hidey2016identifying} \cite{tao2023unievent}.

Sub-event relation extraction: we evaluate \textbf{HiEve} \cite{glavavs2014hieve} and MAVEN-ERE (\textbf{MAVEN-intra-subevent}) \cite{wang2022maven}.

In addition, temporal relation extraction is a multi-classification task (our method only supports binary classification), and also, there are no published studies on temporal relation classification and sub-event relation classification, so we did not evaluate these tasks in the main experiment. Finally, binary-F1 score is used as the main evaluation metric in all tasks.

\begin{table*}
  \centering
  \resizebox{6.2in}{!}{
  \begin{tabular}{lcccccccccc}
    \hline
    \textbf{Model} & \textbf{SCI-uni} & \textbf{ESL-uni} & \textbf{CTB-uni} & \textbf{ESC-intra} & \textbf{CTB-intra} & \textbf{MAVEN-intra-causal} & \textbf{CNC} & \textbf{ALT-uni} & \textbf{MAVEN-intra-subevent} & \textbf{HiEve} \\
    \hline
    T5 & 49.89* & 31.40* & 3.49* & 30.19 & 8.37 & 30.36 & 51.01 & 67.90*  & 6.58 & 12.12 \\
    T5-large & 51.03 & 32.61 & 4.46 & 30.05 & 6.09 & 30.05 & 66.37 & 66.41 & 6.57 & 11.18 \\
    T0-3B & 49.87* & 72.21* & 4.39* & 28.38 & 6.59 & 29.12 & 67.74 & 68.03*  & 6.70 & 10.22 \\
    UniEvent & 82.78*  & 70.64*  & 8.95* & -- & -- & --  & -- & 62.50* & -- & -- \\
    text-davinci-002 & -- & -- & -- & 36.00*  & 9.30* & 32.40* & -- & -- & -- & -- \\
    text-davinci-003 & -- & -- & -- & \textbf{45.90}* & 15.00* & \textbf{37.50}* & -- & -- & -- & -- \\
    Claude-3.5 & 62.60 & 63.60 & 4.91 & -- & -- & -- & 52.70 & 68.18   & 9.59 & 15.83 \\
    GPT-3.5 & 40.99 & 48.56 & 4.71 & 41.00* & 12.80* & 32.30* & 63.10 & 63.57   & 7.85 & 9.90 \\
    GPT-4 & 41.58 & 54.57 & 2.49 & 42.20* & 11.50* & 36.20* & 61.90 & 67.57   & 10.33 & 9.84 \\
    ABLE & \textbf{83.48} & \textbf{72.42} & \textbf{13.64} & 38.48 & \textbf{21.63} & 37.43 & \textbf{69.90} & \textbf{68.42} & \textbf{12.59} & \textbf{17.69} \\
    \hline
  \end{tabular}}
  \caption{Results on zero-shot inter-event causal relation extraction, causal relation classification, and sub-event relation extraction. The best results for each dataset are bolded, * indicates that the original paper results are cited, and the others are the results we reproduced.}
  \label{tab2}
\end{table*}

\begin{table*}
  \centering
  \resizebox{6.2in}{!}{
  \begin{tabular}{lcccccccccccccccccc}
    \hline
    \multirow{2}{*}{\textbf{Model}} & \multicolumn{3}{c}{\textbf{SCI-uni}} & \multicolumn{3}{c}{\textbf{ESL-uni}} & \multicolumn{3}{c}{\textbf{CTB-uni}} & \multicolumn{3}{c}{\textbf{ESC-intra}} & \multicolumn{3}{c}{\textbf{CTB-intra}} & \multicolumn{3}{c}{\textbf{MAVEN-intra}}\\
    \cmidrule(r){2-4} \cmidrule(r){5-7} \cmidrule(r){8-10} \cmidrule(r){11-13} \cmidrule(r){14-16} \cmidrule(r){17-19}
    ~ & P & R & F1 & P & R & F1 & P & R & F1 & P & R & F1 & P & R & F1 & P & R & F1\\
    \hline
    \begin{math} {ABLE}_{Enc}^{1} \end{math} & 67.05 & 79.05 & 72.56 & 83.33 & 64.04 & \textbf{72.42} & 3.12 & 21.05 & 5.44 & 27.95 & 61.75 & \textbf{38.48} & 6.85 & 36.58 & 11.53 & 27.15 & 60.23 & \textbf{37.43} \\
    \begin{math} {ABLE}_{Enc}^{2} \end{math} & 66.94 & 82.09 & 73.75 & 71.04 & 64.04 & 67.36 & 2.67 & 31.58 & 4.92 & 29.74 & 54.52 & \textbf{38.48} & 6.40 & 27.52 & 10.39 & 25.50 & 65.25 & 36.67 \\
    \begin{math} {ABLE}_{Dec}^{1} \end{math} & 78.39 & 82.09 & 80.20 & 74.68 & 56.65 & 64.43 & 6.06 & 10.53 & 7.69 & 30.82 & 49.89 & 38.10 & 13.22 & 26.17 & 17.57 & 22.21 & 74.34 & 34.20 \\
    \begin{math} {ABLE}_{Dec}^{2} \end{math} & 86.02 & 81.08 & \textbf{83.48} & 64.47 & 48.28 & 55.21 & 12.00 & 15.79 & \textbf{13.64} & 25.40 & 68.53 & 37.07 & 16.19 & 32.55 & \textbf{21.63} & 24.57 & 71.60 & 36.58 \\
    \begin{math} {ROLE}_{Enc} \end{math} & 85.15 & 68.18 & 75.73 & 71.60 & 60.10 & 65.35 & 2.37 & 84.21 & 4.62 & 26.52 & 65.71 & 37.78 & 5.71 & 27.85 & 9.48 & 28.11 & 48.59 & 35.62 \\
    \begin{math} {ROLE}_{Dec} \end{math} & 81.11 & 76.57 & 78.78 & 59.31 & 62.69 & 60.96 & 0.78 & 55.56 & 1.54 & 23.82 & 78.79 & 36.58 & 11.99 & 20.81 & 15.21 & 27.01 & 51.53 & 35.44 \\
    \begin{math} w/o All \end{math} & 34.66 & 100.00 & 51.03 & 21.26 & 69.95 & 32.61 & 2.28 & 94.74 & 4.46 & 17.79 & 96.55 & 30.05 & 3.15 & 93.96 & 6.09 & 17.73 & 98.41 & 30.05 \\
    \hline
  \end{tabular}}
  \caption{Ablation experiments on causal relation extraction. The highest F1 scores under each dataset are bolded.}
  \label{tab3}
\end{table*}

\subsection{Baselines}

\textbf{T5} and \textbf{T5-large} \cite{raffel2020exploring}: is a pre-trained language model based on the Transformer architecture, containing encoders and decoders, for a variety of natural language processing tasks.

\textbf{T0-3B} \cite{sanh2022multitask}: a language model optimized for zero-shot learning scenarios based on the T5 architecture.

\textbf{UniEvent} \cite{tao2023unievent}: based on the T5 architecture, utilizes prefix-tuning and multi-task learning to achieve zero-shot event-relational reasoning.

GPT series models \cite{gao2023chatgpt}: including \textbf{text-davinci-002}, \textbf{text-davinci-003}, \textbf{GPT-3.5}, and \textbf{GPT-4}, which are progressively optimized based on OpenAI's GPT-3 model, improving the ability to handle complex tasks. For datasets with no readily available results, we conduct experiments using the API\footnote{\url{https://platform.openai.com}} provided by OpenAI and our designed prompts (see Table~\ref{tab10} in Appendix~\ref{appendix_A}).

\textbf{Claude-3.5 Sonnet}: a language model developed by Anthropic that performs well in zero-shot learning scenarios. We conduct experiments using the APIs\footnote{\url{https://docs.anthropic.com}} provided by Anthropic.

\subsection{Zero-Shot Results}

Table~\ref{tab2} shows the performance of ABLE and the baseline models on the zero-shot causal relation extraction, causal relation classification, and sub-event relation extraction tasks. For causal relation extraction, ABLE achieves SOTA on the SCI-uni, ESL-uni, CTB-uni, and CTB-intra datasets, and shows competitive performance comparable to LLMs on the ESC-intra and MAVEN-intra datasets. For causal relation classification and sub-event relation extraction, ABLE achieves SOTA on all datasets. These results show that ABLE efficiently learns and transfers reasoning knowledge, which improves the performance of various types of zero-shot event-relational reasoning tasks.

\subsection{Ablation study}

To verify the effectiveness of ROLE and ABLE, we conduct ablation experiments. We construct four forms of ABLE, including \begin{math} \bm{{ABLE}_{Enc}^{1}} \end{math}, \begin{math} \bm{{ABLE}_{Enc}^{2}} \end{math}, \begin{math} \bm{{ABLE}_{Dec}^{1}} \end{math}, and \begin{math} \bm{{ABLE}_{Dec}^{2}} \end{math} (see Table~\ref{tab6} for the specific forms); two forms of ROLE, including \begin{math} \bm{{ROLE}_{Enc}} \end{math} and \begin{math} \bm{{ROLE}_{Dec}} \end{math}; and \begin{math} \bm{w/o All} \end{math} (Flan-T5-large without applying ROLE and ABLE). The subscripts Enc and Dec indicate that our method is applied to the encoder and decoder.

Table~\ref{tab3}, Table~\ref{tab4} and Table~\ref{tab5} show the results, which are evaluated by Precision (P), Recall (R) and F1 score (F1). From Table~\ref{tab3} and Table~\ref{tab5}, it is observed that \begin{math} {ROLE}_{Enc} \end{math} and \begin{math} {ROLE}_{Dec} \end{math} improve the precision and F1 score of causal and sub-event relation extraction task, which achieves the goal of ROLE and validates its effectiveness. \begin{math} {ABLE}_{Enc}^{1} \end{math}, \begin{math} {ABLE}_{Enc}^{2} \end{math}, \begin{math} {ABLE}_{Dec}^{1} \end{math}, and \begin{math} {ABLE}_{Dec}^{2} \end{math} improve the F1 score of all tasks , which validates its effectiveness.

Table~\ref{tab4} shows that ROLE performs poorly because the key positions of the positive and negative samples partially overlap in this task, and ROLE strengthens the positive sample to predict ``No'', which leads to a decrease in the recall. ABLE improves the F1 score, demonstrating its strong knowledge transfer capability.

In addition, based on the best results of ABLE in each table, we observe that strong analogies are shown between causal and temporal relations, and between causal and sub-event relations, but the analogies between temporal and sub-event relations are relatively weak (see the results of \begin{math} {ABLE}_{Enc}^{2} \end{math} and \begin{math} {ABLE}_{Dec}^{2} \end{math} in Table~\ref{tab5}).

\begin{table}[H]
  \centering
  \resizebox{2.5in}{!}{
  \begin{tabular}{lcccccc}
    \hline
    \multirow{2}{*}{\textbf{Model}} & \multicolumn{3}{c}{\textbf{ALT}} & \multicolumn{3}{c}{\textbf{CNC}} \\
    \cmidrule(r){2-4} \cmidrule(r){5-7}
    ~ & P & R & F1 & P & R & F1 \\
    \hline
    \begin{math} {ABLE}_{Enc}^{1} \end{math} & 54.64 & 86.03 & 66.83 & 56.96 & 82.09 & 67.26 \\
    \begin{math} {ABLE}_{Enc}^{2} \end{math} & 53.38 & 92.70 & 67.75 & 55.60 & 90.88 & 68.99 \\
    \begin{math} {ABLE}_{Dec}^{1} \end{math} & 52.28 & 98.10 & 68.21 & 53.90 & 98.62 & \textbf{69.70} \\
    \begin{math} {ABLE}_{Dec}^{2} \end{math} & 52.26 & 99.05 & \textbf{68.42} & 53.99 & 99.12 & 69.90 \\
    \begin{math} {ROLE}_{Enc} \end{math} & 55.36 & 82.95 & 66.40 & 57.31 & 78.21 & 66.15 \\
    \begin{math} {ROLE}_{Dec} \end{math} & 55.73 & 81.31 & 66.13 & 57.71 & 72.98 & 64.46 \\
    \begin{math} w/o All \end{math} & 55.22 & 83.28 & 66.41 & 57.11 & 79.22 & 66.37 \\
    \hline
  \end{tabular}}
  \caption{Ablation experiments on causal relation classification. The highest F1 scores under each dataset are bolded.}
  \label{tab4}
\end{table}

\begin{table}[H]
  \centering
  \resizebox{2.5in}{!}{
  \begin{tabular}{lcccccc}
    \hline
    \multirow{2}{*}{\textbf{Model}} & \multicolumn{3}{c}{\textbf{MAVEN-intra}} & \multicolumn{3}{c}{\textbf{HiEve}} \\
    \cmidrule(r){2-4} \cmidrule(r){5-7}
    ~ & P & R & F1 & P & R & F1 \\
    \hline
    \begin{math} {ABLE}_{Enc}^{1} \end{math} & 8.10 & 28.33 & \textbf{12.59} & 10.58 & 43.20 & 16.99 \\
    \begin{math} {ABLE}_{Enc}^{2} \end{math} & 3.41 & 98.18 & 6.59 & 59.00 & 98.82 & 11.14 \\
    \begin{math} {ABLE}_{Dec}^{1} \end{math} & 5.26 & 40.15 & 9.30 & 11.54 & 37.80 & \textbf{17.69} \\
    \begin{math} {ABLE}_{Dec}^{2} \end{math} & 4.01 & 58.94 & 7.51 & 9.70 & 35.69 & 15.25 \\
    \begin{math} {ROLE}_{Enc} \end{math} & 7.35 & 20.46 & 10.82 & 10.72 & 38.79 & 16.79 \\
    \begin{math} {ROLE}_{Dec} \end{math} & 5.75 & 42.00 & 10.11 & 11.33 & 28.34 & 16.19 \\
    \begin{math} w/o All \end{math} & 3.40 & 97.88 & 6.57 & 5.92 & 98.82 & 11.18 \\
    \hline
  \end{tabular}}
  \caption{Ablation experiments on sub-event relation extraction. The highest F1 scores under each dataset are bolded.}
  \label{tab5}
\end{table}

\subsection{Analysis of the analogicality of location}

This subsection analyzes the analogicality of the location of key modules. First, Table~\ref{tab6} shows the layers of editing for ROLE and ABLE under different tasks. \begin{math} {ROLE}_{Enc} \end{math} selects the top 3 module layers in terms of average indirect effects (see Equation~\ref{eq2}) in negative samples. \begin{math} {ROLE}_{Dec} \end{math} selects the 1st ranked module layer (the sub-event relation extraction task selects the 3rd ranked layer because the 1st and 2nd ranked positions overlap for the positive and negative samples). ABLE determines the module layer analogously (see Equation~\ref{eq8}). Table~\ref{tab6} shows that ABLE obtains positions that are close to the top ranked positions obtained by ROLE (see Table~\ref{tab11} in Appendix~\ref{appendix_B}), which validates the analogicality of the location to some extent.

\begin{figure*}[t]
  \begin{center}
    \includegraphics[width=0.6\linewidth]{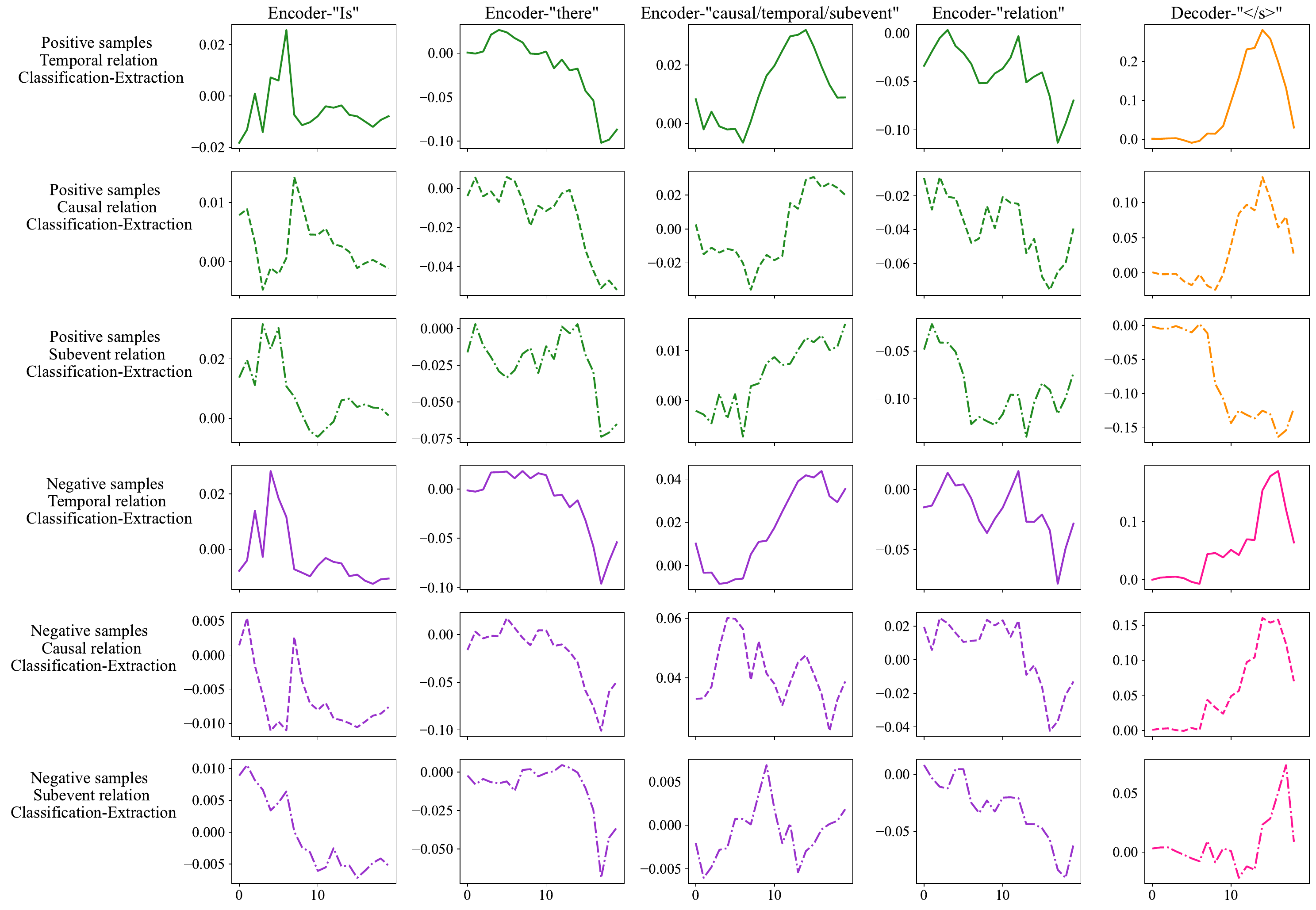}
    \caption{Line plots of the effect difference between classification and extraction for temporal, causal, and subevent relations. Horizontal coordinates indicate the layer of the module and vertical coordinates indicate the difference in average indirect effects. The first three rows indicate positive samples and the last three rows indicate negative samples. The first four columns show the four tokens of the encoder and the last column shows the “</s>” token of the decoder.}
    \label{Fig6}
  \end{center}
\end{figure*}

Second, as seen from Figure~\ref{Fig6}, the difference line plots of temporal, causal, and sub-event relations show similar trends for either positive or negative samples for most tokens, which further validates the analogous nature of location.

Additionally, for the positive samples, in the decoder's “</s>” token (first 3 rows of the last column), the line plots of the causal and temporal relations are similar, while the plots of the sub-event relations are different. For the negative samples, in the encoder's “causal/temporal/sub-event” token (the last 3 rows of the 3rd column), the line plots of causal and subevent relations are similar, while the plots of temporal relations are different. These results indicate a strong analogical nature between the causal and temporal relations, as well as between the causal and sub-event relations, and a weaker analogous nature between the temporal and sub-event relations. This also explains the limited effect of \begin{math} {ABLE}_{Enc}^{2} \end{math} and \begin{math} {ABLE}_{Dec}^{2} \end{math} in Table~\ref{tab5}.

\begin{table}
  \centering
  \resizebox{3in}{!}{
  \begin{tabular}{lp{4cm}p{4cm}p{4cm}}
    \hline
    \textbf{\textbf{Model}} & \textbf{Causal relation extraction} & \textbf{Causal relation classification} & \textbf{Sub-event relation extraction} \\
    \hline
    \begin{math} {ROLE}_{Enc} \end{math} & [15,16,17] & [15,16,17] & [16,17,18] \\
    \hline
    \begin{math} {ROLE}_{Dec} \end{math} & [17] & [14] & [15] \\
    \hline
    \begin{math} {ABLE}_{Enc}^{1} \end{math} 
     & \makecell[l]{Sub-event cla: [17] \\ Sub-event ext: [17] \\ Causal cla: [17] \\ \begin{math} \rightarrow \end{math} \textbf{Causal ext: [17]}}
     & \makecell[l]{Sub-event ext: [17] \\ Sub-event cla: [17] \\ Causal ext: [17] \\ \begin{math} \rightarrow \end{math} \textbf{Causal cla: [17]}}
     & \makecell[l]{Causal cla: [17] \\ Causal ext: [17] \\ Sub-event cla: [17] \\ \begin{math} \rightarrow \end{math} \textbf{Sub-event ext: [17]}} \\
    \hline
    \begin{math} {ABLE}_{Enc}^{2} \end{math}
     & \makecell[l]{Temporal cla: [15,16,17] \\ Temporal ext: [16,17,18] \\ Causal cla:  [15,16,17] \\ \begin{math} \rightarrow \end{math} \textbf{Causal ext: [16,17,18]}}
     & \makecell[l]{Temporal ext: [16,17,18] \\ Temporal cla: [15,16,17] \\ Causal ext: [15,16,17] \\ \begin{math} \rightarrow \end{math} \textbf{Causal cla: [14,15,16]}}
     & \makecell[l]{Temporal cla: [16] \\ Temporal ext: [18] \\ Sub-event cla: [17] \\ \begin{math} \rightarrow \end{math} \textbf{Sub-event ext: [19]}} \\
    \hline
    \begin{math} {ABLE}_{Dec}^{1} \end{math} 
     & \makecell[l]{Sub-event cla: [17] \\ Sub-event ext: [15] \\ Causal cla: [14] \\ \begin{math} \rightarrow \end{math} \textbf{Causal ext: [12]}}
     & \makecell[l]{Sub-event ext: [15] \\ Sub-event cla: [17] \\ Causal ext: [12] \\ \begin{math} \rightarrow \end{math} \textbf{Causal cla: [14]}}
     & \makecell[l]{Causal cla: [14] \\ Causal ext: [12] \\ Sub-event cla: [17] \\ \begin{math} \rightarrow \end{math} \textbf{Sub-event ext: [15]}} \\
    \hline
    \begin{math} {ABLE}_{Dec}^{2} \end{math}
     & \makecell[l]{Temporal cla: [15] \\ Temporal ext: [13] \\ Causal cla: [14] \\ \begin{math} \rightarrow \end{math} \textbf{Causal ext: [12]}}
     & \makecell[l]{Temporal ext: [11,12,13] \\ Temporal cla: [14,15,16] \\ Causal ext:  [11,12,13] \\ \begin{math} \rightarrow \end{math} \textbf{Causal cla: [14,15,16]}}
     & \makecell[l]{Temporal cla: [14,15,16] \\ Temporal ext: [11,12,13] \\ Sub-event cla:  [14,15,16] \\ \begin{math} \rightarrow \end{math} \textbf{Sub-event ext: [11,12,13]}} \\
    \hline
  \end{tabular}}
  \caption{The layers of editing for different models under each task, and the analogous forms of \begin{math} {ABLE}_{Enc}^{1} \end{math}, \begin{math} {ABLE}_{Enc}^{2} \end{math}, \begin{math} {ABLE}_{Dec}^{1} \end{math}, and \begin{math} {ABLE}_{Dec}^{2} \end{math}. The subscripts Enc and Dec indicate that our method is applied to encoder and decoder respectively.}
  \label{tab6}
\end{table}

\subsection{Analysis of the analogicality of editing magnitude}

This subsection analyzes the analogicality of editing magnitude. Let \begin{math} A \end{math}, \begin{math} B \end{math}, \begin{math} C \end{math}, \begin{math} D \end{math}, \begin{math} E \end{math}, \begin{math} F \end{math} denote the classification and extraction tasks of temporal, causal, and sub-event relations, respectively. Let \begin{math} {\mathrm{\Delta}W}_{X} \end{math} denote the editing magnitude for a module parameter of task \begin{math} 
X \end{math}, and let \begin{math} {\mathrm{\Delta}W}_{XY} = {\mathrm{\Delta}W}_{X} - {\mathrm{\Delta}W}_{Y} \end{math}, where \begin{math} 
X \in \left\{ {A,C,E} \right\} \end{math} and \begin{math} 
Y \in \left\{ {B,D,F} \right\} \end{math}.

Table~\ref{tab7} shows the similarity between different \begin{math} {\mathrm{\Delta}W}_{XY} \end{math}. We use the cosine of the main eigenvectors of the matrices to compute the similarity, because it reflects the degree of similarity between the main transformation directions of the matrices. If the similarity is near 1, the directions are similar; if it is near -1, the directions are opposite.

As seen in Table~\ref{tab7}, similarities between \begin{math} {\mathrm{\Delta}W}_{XY} \end{math} of analogous tasks are higher than those between \begin{math} {\mathrm{\Delta}W}_{XY} \end{math} of non-analogous tasks, which verifies the analogicality of editing magnitude to some extent.

\begin{table}
  \centering
  \resizebox{3in}{!}{
  \begin{tabular}{llllll}
    \hline
    & Analogous tasks & Non-analogous tasks & Non-analogous tasks \\
    \hline
    & \begin{math} sim\left( {{\mathrm{\Delta}W}_{AB},{\mathrm{\Delta}W}_{CD}} \right) \end{math}              & \begin{math} sim\left( {{\mathrm{\Delta}W}_{AB},{\mathrm{\Delta}W}_{CF}} \right) \end{math}               & \begin{math} sim\left( {{\mathrm{\Delta}W}_{AB},{\mathrm{\Delta}W}_{AD}} \right) \end{math}               \\
    \hline
    Sim & \textbf{0.08}            & 0.06             & -0.05            \\
    \hline
    & \begin{math} sim\left( {{\mathrm{\Delta}W}_{CD},{\mathrm{\Delta}W}_{EF}} \right) \end{math}              & \begin{math} sim\left( {{\mathrm{\Delta}W}_{CD},{\mathrm{\Delta}W}_{EB}} \right) \end{math}               & \begin{math} sim\left( {{\mathrm{\Delta}W}_{CD},{\mathrm{\Delta}W}_{CF}} \right) \end{math}               \\
    \hline
    Sim & \textbf{0.19}            & -0.04            & -0.20            \\
    \hline
    & \begin{math} sim\left( {{\mathrm{\Delta}W}_{EF},{\mathrm{\Delta}W}_{AB}} \right) \end{math}              & \begin{math} sim\left( {{\mathrm{\Delta}W}_{EF},{\mathrm{\Delta}W}_{AD}} \right) \end{math}               & \begin{math} sim\left( {{\mathrm{\Delta}W}_{EF},{\mathrm{\Delta}W}_{EB}} \right) \end{math}               \\
    \hline
    Sim & \textbf{0.05}            & 0.04             & -0.03             \\
    \hline
  \end{tabular}}
  \caption{Similarity between different \begin{math} {\mathrm{\Delta}W}_{XY} \end{math}. “Sim” denotes the cosine similarity of the main eigenvectors of the matrix.}
  \label{tab7}
\end{table}

\subsection{Analysis on computational resources and time}

We analyze the computational resources and time required for ROLE, ABLE, the fine-tuning method and UniEvent (prefix fine-tuning), respectively, with respect to the amount of parameters (Params), GPU memory consumption and time for model training, as shown in Table~\ref{tab8}. We show the results of ROLE and ABLE for encoder (Enc) and decoder (Dec). To facilitate the implementation of UniEvent, we add prefixes to the decoder and set the length to 5. All models use 10 training samples with a training epoch of 10 and a batch size of 1.

Table~\ref{tab8} shows that \begin{math} {ROLE}_{Dec} \end{math} and \begin{math} {ABLE}_{Dec} \end{math} consume less computational cost than \begin{math} {ROLE}_{Enc} \end{math} and \begin{math} {ABLE}_{Enc} \end{math} since fewer parameters are edited in the decoder. Moreover, our method is significantly lower than the fine-tuning method and UniEvent in all the metrics, which verifies the efficiency of our method.

\begin{table}
\centering
\resizebox{3in}{!}{
\begin{tabular}{lccc}
\hline
                             & Params (M) & GPU memory (MB) & Training time (s) \\
\hline
\begin{math} {ROLE}_{Enc}/{ROLE}_{Dec} \end{math} 
                             & 2.88 / 1.05 & 3969 / 3789 & 12.60 / 8.60              \\
\begin{math} {ABLE}_{Enc}/{ABLE}_{Dec} \end{math} 
                             & 11.53 / 4.19 & 3761 / 3729 & 0.09 / 0.03              \\
UniEvent                     & 50.95      & 6735            & 24.53             \\
Fine-tuning                  & 783.15     & 13951           & 23.23              \\ 
\hline
\end{tabular}}
\caption{Computational resources and time required for ROLE, ABLE, fine-tuning methods and UniEvent on the CTB-uni dataset, respectively.}
\label{tab8}
\end{table}

\section{Conclusion}

We first propose ROLE to locate and edit key modules of language models in reasoning about event relations. The results show that ROLE improves interpretability and reasoning performance with reduced computational cost. Then, we propose ABLE to analogize the similarities and differences between tasks. The results show that ABLE achieves SOTA results for zero-shot event-relational reasoning on most datasets.

Furthermore, our experiments provide insights into the mechanisms of reasoning about event relations in language models and verify the feasibility of model editing to optimize reasoning capabilities. Future work could utilize ROLE and ABLE to further explore the reasoning capabilities of large language models.

\section{Limitations}

Our methods mainly address binary reasoning tasks. Moreover, our study on the reasoning ability of language models is not comprehensive enough. Therefore, future work can be extended to more complex reasoning scenarios, and also, more experiments can be conducted to explore the reasoning mechanism in depth.

\section*{Acknowledgments}

This work is supported by grant from the National Natural Science Foundation of China (No. 62076048), the Science and Technology Innovation Foundation of Dalian (2020JJ26GX035).

\bibliography{custom}

\appendix

\section{Data Preprocessing in Reasoning-Oriented Locating}
\label{appendix_A}

We use the MAVEN dataset \cite{wang2022maven} for reasoning-oriented locating, which contains temporal, causal, and sub-event relations, and its statistical information is shown in Table~\ref{tab9}. To explore the key modules of the language model in reasoning about event relations, we design 6 tasks, including classification and extraction tasks for temporal, causal and sub-event relations. Extraction tasks aim to identify event relations given context and head and tail events, and classification tasks aim to identify event relations given context only.

Subsequently, we designed prompt and verbalizer (see Table~\ref{tab10}) for each task to stimulate the reasoning ability of the language model. Meanwhile, for each task, we randomly selected 500 positive and 500 negative samples for analysis. To make the key positions of each task centralized, we pick only one class of relations as the positive class. Specifically:

(1) For temporal relations, we only selected the samples of “BEFORE” relations as positive samples, and the samples other than “BEFORE, OVERLAP, CONTAINS, SIMULTANEOUS, ENDS-ON, BEGINS-ON” as negative samples.

(2) For causal relations, we selected the samples of “CAUSE” relations as positive samples, and the samples other than “CAUSE, PRECONDITION” as negative samples.

(3) For sub-event relations, we directly selected the samples of “SUBEVENT” relation as positive samples, and the samples other than “SUBEVENT” as negative samples.

\section{Average Indirect Effects Ranking of Layers for Specific Modules in Various Tasks}
\label{appendix_B}

Table~\ref{tab11} and Table~\ref{tab12} show the layer ordering (in ascending order) of the average indirect effects for the encoder's MLP module and the decoder's cross-attention module. We focus on the top 3 ranked layers of indirect effects in each task.

\begin{table*}
  \centering
  \resizebox{5in}{!}{
  \begin{tabular}{lcp{7cm}}
    \hline
    \textbf{Relation} & \textbf{The number of samples} & \textbf{Specific relation types} \\
    \hline
    Temporal relation & 1216217 & BEFORE, OVERLAP, CONTAINS, SIMULTANEOUS, ENDS-ON, BEGINS-ON \\
    Causal relation & 57992 & CAUSE, PRECONDITION \\
    Sub-event relation & 15841 &  \\
    \hline
  \end{tabular}}
  \caption{Statistical information on temporal, causal and sub-event relations in the MAVEN dataset.}
  \label{tab9}
\end{table*}

\begin{table*}
  \centering
  \resizebox{6in}{!}{
  \begin{tabular}{lp{6cm}l}
    \hline
    \textbf{Task} & \textbf{Prompt} & \textbf{Verbalizer} \\
    \hline
    Temporal relation classification & Answering the following yes/no question. Is there a temporal relation in <Sentence>? & TEMPORAL: Yes, NONE: No \\
    Temporal relation extraction & Answering the following yes/no question. Is there a temporal relation between <event1> and <event2> in <Sentence>? & TEMPORAL: Yes, NONE: No \\
    Causal relation classification & Answering the following yes/no question. Is there a causal relation in <Sentence>? & CAUSAL: Yes, NONE: No \\
   Causal relation extraction & Answering the following yes/no question. Is there a causal relation between <event1> and <event2> in <Sentence>? & CAUSAL: Yes, NONE: No \\
    Sub-event relation classification & Answering the following yes/no question. Is there a sub-event relation in <Sentence>? & SUB-EVENT: Yes, NONE: No \\
    Sub-event relation extraction & Answering the following yes/no question. Is there a sub-event relation between <event1> and <event2> in <Sentence>? & SUB-EVENT: Yes, NONE: No \\
    \hline
  \end{tabular}}
  \caption{Prompts and verbalizers for classification and extraction tasks of temporal, causal and sub-event relations. Extraction tasks aim to identify event relations given context and head and tail events, and classification tasks aim to identify event relations given context only.}
  \label{tab10}
\end{table*}

\begin{table*}
  \centering
  \resizebox{6.2in}{!}{
  \begin{tabular}{p{5cm}p{7cm}p{7cm}}
    \hline
    \textbf{Task} & \textbf{MLP module in encoder for “Is” token} & \textbf{Cross-Attention module in decoder for “\textless{}s\textgreater{}” token} \\
    \hline
    Temporal relation classification & 1, 0, 4, 3, 5, 2, 8, 6, 7, 19, 13, 11, 14, 12, 9, 10, 18, 15, 17, 16 & 5, 7, 0, 3, 1, 2, 4, 6, 8, 9, 18, 10, 11, 12, 13, 17, 16, 14, 15 \\
    Temporal relation extraction & 0, 1, 4, 3, 5, 6, 2, 8, 7, 9, 13, 14, 12, 11, 10, 19, 15, 16, 17, 18 & 7, 3, 2, 1, 5, 0, 4, 8, 6, 9 ,18, 16, 10, 15, 17, 14, 11, 12, 13 \\
    Causal relation classification & 0, 1, 8, 19, 2, 13, 14, 4. 5, 3, 7, 11, 10, 6, 12, 9, 18, 16, 15, 17 & 7, 3, 0, 2, 1, 4, 5, 6, 8, 9, 10, 18, 11, 12, 16, 13, 15, 17, 14 \\
   Causal relation extraction & 0, 1, 2, 8, 4, 19, 3, 5, 6, 13 ,14, 7, 10, 11, 9, 12, 18, 16, 15, 17 & 7, 2, 0, 3, 1, 4, 5, 8, 6, 9, 16, 10, 18, 15, 14, 13, 12, 11, 17 \\
    Sub-event relation classification & 1, 4, 3, 2, 0, 5, 8, 6, 7, 9, 11, 13, 19, 10, 14, 12, 15, 16, 18, 17 & 7, 5, 4, 3, 6, 0, 2, 1, 8, 9, 10, 18, 11, 12, 13, 16, 15, 14, 17 \\
    Sub-event relation extraction & 1, 4, 0, 3, 5, 2, 8, 6, 7, 9, 10, 11, 13, 14, 12, 19, 15, 16, 18, 17 & 7, 2, 3, 1, 0, 5, 4, 6, 8, 9, 10, 18, 11, 16, 12, 13, 15, 14, 17 \\
    \hline
  \end{tabular}}
  \caption{Layer ordering of average indirect effects of specific modules on specific tokens for \textbf{negative} samples in each task.}
  \label{tab11}
\end{table*}

\begin{table*}
\centering
\resizebox{6.2in}{!}{
\begin{tabular}{p{5cm}p{5cm}p{5cm}p{5cm}}
\hline
\textbf{Task} & \textbf{MLP module in encoder for “relation” token} & \textbf{MLP module in encoder for “relation” token} & \textbf{Cross-Attention module in decoder for “\textless{}s\textgreater{}” token} \\
\hline
Temporal relation classification  & 11, 12, 13, 10, 14, 15, 3, 1, 7, 6, 4, 9, 2, 8, 0, 5, 16, 19, 18, 17 & 15, 14, 13, 8, 16, 9, 12, 10, 0, 2, 11, 7, 4, 6, 1, 3, 17, 5, 18, 19 & 1, 0, 2, 3, 4, 5, 6, 7, 8, 9, 18, 10, 11, 13, 12, 17, 16, 15, 14         \\
Temporal relation extraction      & 12, 11, 13, 14, 10, 15, 9, 8, 0, 2, 7, 3, 1, 4, 6, 5, 16, 19, 18, 17 & 12, 2, 3, 15, 4, 11, 14, 0, 1, 10, 9, 13, 5, 8, 6, 16, 7, 19, 18, 17 & 2, 1, 0, 3, 4, 5, 7, 6, 8, 13, 12, 9, 11, 10, 14, 15, 18, 16, 17         \\
Causal relation classification    & 7, 11, 8, 12, 13, 10, 14, 9, 0, 1, 15, 2, 3, 6, 5, 4, 16, 19, 17, 18 & 15, 12, 13, 5, 1, 4, 9, 14, 16, 6, 0, 7, 10, 3, 11, 8, 17, 2, 18, 19 & 0, 2, 1, 3, 4, 5, 7, 6, 8, 9, 13, 10, 12, 11, 18, 15, 14, 17, 16         \\
Causal relation extraction        & 14, 15, 12, 13, 0, 11, 8, 9, 10, 1, 7, 2, 3, 4, 5, 16, 6, 17, 19, 18 & 12, 0, 4, 1, 5, 10, 2, 3, 9, 15, 11, 14, 8, 13, 6, 7, 16, 19, 17, 18 & 0, 2, 3, 1, 4, 6, 5, 7, 13, 12, 11, 8, 10, 9, 14, 15, 17, 18, 16         \\
Sub-event relation classification & 17, 2, 3, 1, 18, 4, 12, 19, 6, 16, 0, 11, 5, 9, 10, 8, 7, 15, 14, 13 & 6, 9, 10, 8, 13, 0, 11, 7, 12, 14, 5, 4, 2, 15, 1, 3, 16, 17, 18, 19 & 2, 1, 0, 3, 4, 7, 5, 6, 8, 9, 10, 13, 11, 12, 18, 15, 14, 16, 17         \\
Sub-event relation extraction     & 17, 19, 18, 16, 12, 3, 11, 1, 2, 14, 10, 15, 9, 4, 0, 13, 8, 5, 7, 6 & 0, 1, 2, 4, 3, 5, 11, 6, 10, 12, 9, 15, 14, 8, 7, 16, 13, 19, 17, 18 & 0, 2, 1, 3, 4, 7, 5, 6, 8, 9, 11, 13, 12, 10, 18, 15, 14, 16, 17          \\
\hline
\end{tabular}}
\caption{Layer ordering of average indirect effects of specific modules on specific tokens for \textbf{positive} samples in each task.}
\label{tab12}
\end{table*}

\section{Prompt Segmentation Algorithm Based on Singular Value Decomposition}
\label{appendix_C}

We propose an algorithm based on Singular Value Decomposition (SVD) for segmenting the tokens in the prompt as shown in Table~\ref{Algorithm1}. The algorithm aims to utilize the linear independence between the indirect effect vectors of neighboring tokens to determine the division point.

Specifically, the initial state is set to 2 eigenvalues. As the new token is added, the number of eigenvalues of the indirect effect matrix increases gradually. When the number of eigenvalues increases, it indicates that the relationship between the new token and the existing tokens has changed significantly, and then the token sequence is divided. The steps above are repeated until all the tokens are processed.

\begin{figure*}[t]
  \centering
  \includegraphics[width=0.8\linewidth]{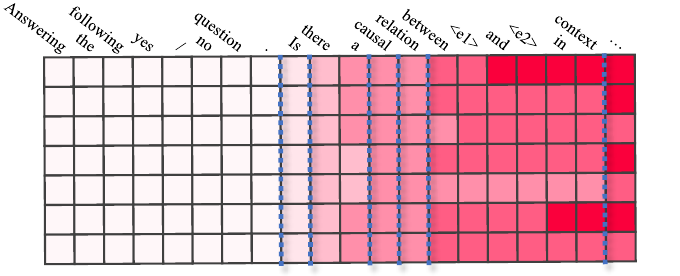}
  \caption{ A graphical illustration of the segmentation results of the SVD-based prompt segmentation algorithm on different samples. Each row represents a sample and each column corresponds to a token. The change in color shade reflects the change in the number of eigenvalues and also indicates the location of segmentation. The blue dotted line indicates the final division.}
  \label{Fig7}
\end{figure*}

Figure~\ref{Fig7} shows the segmentation results of the algorithm on multiple samples. Each row of the figure represents a sample, and each column corresponds to a token. Changes in color shades indicate changes in the number of eigenvalues, and also indicate where the segmentation needs to be performed.

Based on the segmentation results of these samples, we divide the prompts into the following sections: \emph{``Answering the following yes$\backslash$no question.''}, \emph{``Is''}, \emph{``there a''}, \emph{``temporal$\backslash$causal$\backslash$sub-event''}, \emph{``relation''}, \emph{``(between <event1> and <event2>) in sentence''}, \emph{```<context>' ?''}. This division can show the influence of each module on the final result more clearly, which effectively improves the interpretability of the analysis.

\begin{table*}
  \centering
  \begin{tabular}{p{15cm}}
    \hline
    \textbf{Algorithm 1} SVD-Based Prompt Segmentation Algorithm\\
    \hline
    \textbf{Input}: Indirect effect matrix \begin{math} M \end{math} of all tokens under each layer of the module, the length of the token is \begin{math} N \end{math}, and the initialized index \begin{math} i=0 \end{math}. \\
    \textbf{for} \begin{math} i \end{math} in range(\begin{math} N \end{math}): \\
    \begin{math} count \left[i\right] = SVD\left(M \left[: i\right]\right) \end{math} \# Calculate the number of eigenvalues of the current matrix \\
   \textbf{if} \begin{math} count \left[i\right] > count \left[i - 1\right] \end{math}: \\
    \textbf{print} \begin{math} i , \textrm{ } count \left[i\right] \end{math}) \# Show segmentation results \\
    \hline
  \end{tabular}
  \caption{SVD-Based Prompt Segmentation Algorithm}
  \label{Algorithm1}
\end{table*}

\end{document}